\documentclass[journal]{IEEEtran}
\usepackage{eurosym}

\usepackage[bottom]{footmisc}
\usepackage{stfloats}

\usepackage{cite}

\usepackage{enumitem}

\ifCLASSINFOpdf
  \usepackage[pdftex]{graphicx}
  \usepackage{import}
\else
\fi
\usepackage[english]{babel}
\usepackage[utf8]{inputenc}
\usepackage{subcaption}
\usepackage{placeins}
\usepackage{color}
\usepackage{mathtools}
\usepackage{placeins}
\usepackage{relsize}
\usepackage{amssymb}
\usepackage{units}
\usepackage{amsmath}
\usepackage{float} 		

\usepackage{pgf}
\usepackage{tikz}
\usetikzlibrary{arrows,automata}
\usetikzlibrary{positioning}
\tikzset{    state2/.style={ rectangle,  draw=black, inner sep=7pt,  text centered  },}

\usepackage[color=black,opacity=1,angle=0,scale=1]{background}
\backgroundsetup{
contents={
\begin{tikzpicture}
\node at (current page.center) [align=center] {\textcopyright 2019 IEEE. Personal use of this material is permitted. 
\\ Permission from IEEE must be obtained for all other uses, in any current or future media, including reprinting/republishing this material for advertising or promotional purposes, 
\\ creating new collective works, for resale or redistribution to servers or lists, or reuse of any copyrighted component of this work in other works.
\\ DOI: 10.1109/TIV.2019.2919465};
\end{tikzpicture}},
placement=bottom,
scale=0.6,
vshift=20
}

\begin{document}
%
\title{
Probabilistic Uncertainty-Aware Risk Spot \\ Detector for Naturalistic Driving}


\author{Tim Puphal, Malte Probst and Julian Eggert
\vspace{-0.4cm}
\thanks{\hspace{-0.3cm} The authors are with the Honda Research Institute (HRI) Europe, Carl-Legien-Str. 30, 63073 Offenbach, Germany 
	(e-mail: tim.puphal@honda-ri.de; malte.probst@honda-ri.de; julian.eggert@honda-ri.de) %
}}

\maketitle

\begin{abstract}
Risk assessment is a central element for the development and validation of Autonomous Vehicles (AV). It comprises a combination of occurrence probability and severity of future critical events. 
Time Headway (TH) as well as Time-To-Contact (TTC) are commonly used risk metrics and have qualitative relations to occurrence probability. However, they lack theoretical derivations and additionally they are designed to only cover special types of traffic scenarios (e.g.~longitudinal following between single car pairs).
In this paper, we present a probabilistic situation risk model based on survival analysis considerations and extend it to naturally incorporate sensory, temporal and behavioral uncertainties as they arise in real-world scenarios. The resulting Risk Spot Detector (RSD) is applied and tested on naturalistic driving data of a multi-lane boulevard with several intersections, enabling the visualization of road criticality maps. Compared to TH and TTC, our approach is more selective and specific in predicting risk. RSD concentrates on driving sections of high vehicle density where large accelerations and decelerations or approaches with high velocity occur.   
\end{abstract}

\IEEEpeerreviewmaketitle


\vspace{-0.38cm}

\vspace{0.2cm}
\section{Introduction}

\IEEEPARstart{C}{ritical} events are extremely sparse in field operational tests for Autonomous Vehicles (AV). As an example, about 0.1-5 accidents happen per 1 million km on german roads \cite{analysisgroup2018}. 
In order for an AV to have the same collision/km rate with an evidence probability of 50\%, it needs to be 2 times better than the human driver on a 10 million km distance. Consequently, an unfeasible vast amount of real-world data is necessary to validate its safety \cite{winner2018}. 
Since there is no travelling without risk, AV's should additionally be able to transparently show the reasoning behind their actions. Compared to previous mobility technologies, the launch of AV's faces disproportionately high requirements. This validation trap could be circumvented with anf objective traffic situation risk measure, which allows to continuously quantify the overall driving performance.

In technical terms, risk is defined as the occurence probability of a loss multiplied with its consequence or severity. The consideration of the future poses thereby two challenges for risk assessment. On one hand, the involved processes which cause criticalities are inherently uncertain. AV's encounter non-linearities (behavior interaction and feedback loops), sensor inaccuracies (false positive or false negative detections), unknown environment parameters (occlusions or missing map data) as well as unobservable facts (drivers' state of mind) \cite{eggert2018}.  
On the other hand, the variability of traffic situations leads to multiple dimensions in the probability of dangerous events:
\begin{enumerate}[leftmargin=0.4cm]
 \item Different types of risks might arise during the scene progress. Typical risks are vehicle-to-vehicle collision, loss of control in a curve or rule violation. 
 \item Possible subset of entities that are involved in critical events. For collision risks, this implies the pairwise consideration of traffic participants. 
 \item For a particular risk type and subset of concerned entities, multiple evolutions of the scene (e.g. turn left, go straight, turn right or lane change) create distinct events. 
 \item For a particular scene, a countless number of critical events can happen at various predicted states (position, velocity, acceleration, etc.). 
\end{enumerate}

Traditional approaches are based e.g.~on deterministic Time Headway (TH) \cite{transpres2010} and Time-To-Collision (TTC) \cite{horst91} as the most prevalent risk indicators for AV, but do not address the listed issues.  
For this reason, we propose a probabilistic risk estimation method called Risk Spot Detector (RSD) that is able to evaluate general collision risks for all surrounding cars under measurement, prediction, historical and behavior uncertainties. In RSD, we combine a Gaussian method for an instantaneous collision probability with the survival analysis \cite{eggert2014} to retrieve an accumulated critical event probability. The framework of RSD was presented previously in \cite{puphal2018}. In this paper, we extend its functionality to account for low but existing risk in normal driving on multi-lane segments and intersections with dense traffic. The resulting performance from RSD is tested on the naturalistic dataset NGSIM (Next Generation Simulation) \cite{ngsim2006} to visualize experienced criticality levels on road maps. In contrast to TH and TTC, our RSD classifies different hazards in all velocity intervals more selectively and with greater precision. 

The next Section \ref{sec:related} introduces state of the art time-based, probabilistic and ex-post risk metrics. After explaining the basics of RSD in Section \ref{secsec:framework}, we detail in Section \ref{secsec:extensions} our modeling of real-world uncertainties. Section \ref{sec:behavioruncertainty} outlines the difference in behavior extrapolation between RSD, TH as well as TTC and Section \ref{sec:experiments} shows a comparative analysis on NGSIM to find out risky areas. In Section \ref{sec:outlook}, we conclude with a summary and prospect for future research topics.

\subsection{Related Work}
\label{sec:related}

The major fields engaging in risk assessment and visualization are the automotive industry [4-12,14,16-18,21-22], robotics [15,19], aviation [20,23,26] plus aerospace technology [13], civil engineering [24], data science [25] and economics [27]. 
Hereby, risk metrics frequently serve in a cost function for motion planning and find hazards on which an entity has to react. 

While TH and TTC assume and describe the remaining time to a collision event with kinematics (distance and velocity) for longitudinal traffic scenarios, Post-Encroachment Time (PET) \cite{allen78} and 2D-TTC \cite{Damerow2014Risk} generalize this notion to intersections. To even consider kinematical constraints of the acting vehicles, Time-To-Brake (TTB) and Time-To-Steer (TTS) analyze the time until an emergency brake or steering maneuver still succesfully avoids the longitudinal crash \cite{hillenbrand2005}. Because of the simple calculation and intuitive interpretation of TH and TTC, the series product assistance function Adaptive Cruise Control (ACC) controls TH and Collision Mitigation Systems (CMS) often take TTC into account. ACC has thereby been formally verified on highways to be stable and safe for several cars with distributed control \cite{loos2011}.   

Alongside time-based indicators, 
probabilistic risks are able to incorporate uncertainties in the prediction. 
Gaussian methods, such as \cite{garmier2009}, model predicted trajectories of moving entities with spatial normal distributions and calculate their overlap as a collision probability. 
In the process, the Gaussian parameters can also be taken from the covariance matrix of a Kalman Filter. With this in mind, \cite{houenou2014} extrapolates Constant Yaw Rate and Acceleration (CYRA) and \cite{hennes2016} employs a Particle Filter with Dirac delta function as a distribution instead. 
For vehicles, planes and satellites, it is easy to constrain their movement along fixed paths. In contrast, mobile robots and pedestrians have very variable actions in space \cite{kern2011}. To determine the common measurement uncertainties of cars, \cite{chang2017} inspected the errors in position, velocity, heading and acceleration from real sensors and found parameter values for the describing multi-modal Gaussians.

Opposed to calculating an analytical solution for event probabilities, Monte Carlo strategies \cite{schreier2014} are wide-spread in research. They approximate risks by sampling position sequences from distribution functions and comparing the number of collisions with misses. This is especially useful when the direct solution is complex, but requires high computational costs for reliable estimates. For example, \cite{alain2008} apply Monte Carlo for simulating collisions of wheeled robots with specific shapes. Similarly, \cite{schmerling2017} improve the convergence time by importance sampling in three-dimensional ranges of motion. 

If the future behavior is known or easier to estimate, collisions can be checked discretely as well. In one work, traffic participants are projected onto map data paths and predicted longitudinally using prior knowledge (e.g.~stopping at a stop line or driving at constant velocity) \cite{ferguson2008}. Afterwards, a possible crash is determined with intersection checks of geometrical shapes around the assumed positions. Closely related, \cite{christopher2009} detect deviations from assumed paths with a Gaussian process as well as intentions (e.g.~brake or accelerate) via Hidden Markov Models (HMM) and 
\cite{hejase2017} create in the context of Dynamic Probabilistic Risk Assessment (DPRA) discrete cells in spatiotemporal state-space to backtrace faulty maneuvers or system behaviors.

Finally, warning systems compare measured risk values with predefined safety thresholds in real-time to support the driver and critical situations are analyzable after they happened with ex-post scores for the purpose of e.g.~traffic flow management. 
Approaches in this direction include kriging techniques \cite{thakali2015}, which allow to create incident heat maps by extrapolating accident ratios from road locations with data to segments where no data is available. 
Similarly, \cite{itoh2015} derive caution spots on map data from recorded driven acceleration and jerk profiles of cars, and airplane risk zones are constructed with the probability of one point colliding into another volume using the rice formula in \cite{nguyen2018}. Furthermore, the authors of \cite{zou2017} optimized the structure and parameters of a Bayesian network from accident data causes (weather, inattention, lane change, etc.). 

The proposed Risk Spot Detector (RSD) falls into the category of probabilistic risk measures. However, RSD is grounded on a mathematical theory of sparse events. Spatial Gaussian probability distributions feed an inhomogenous Poisson process for a survival analysis. Related work employs Gaussian methods or the survival analysis separately. Aside from that, the Gaussian method is refined by modeling 2D Gaussians which have velocity-dependent growth and bent along the future path. RSD thus focuses not only on uncertainties in position, but also in velocity and predicted time of multiple interacting traffic participants. When parametrizing RSD with real data from NGSIM, we can infer plausible output thresholds for distinct criticality levels. As a result, RSD can more reliably distinguish safe from dangerous situation evolutions of multiple types and is seen as suitable for the validation of AV's.

\section{Risk Spot Detector}
\label{sec:riskdetector}

\subsection{Gaussian Method with Survival Analysis}
\label{secsec:framework}

We start by considering a dynamic driving situation with two traffic participants $\mbox{TP}1$ and  $\mbox{TP}2$ at an arbitrary moment in time $t$. From $t$ on, the target of RSD is to estimate the risk of a critical event that could happen at a future time $t\hspace{-0.0005cm}+\hspace{-0.0005cm}s$, that is, at a temporal distance $s$ into the future. We assume the events to be disruptive and to have no duration. Since most of the commonly used risk measures do not address severity explicitly, we will, for simplicity, concentrate on risk as an event occurrence probability with equal severity events. Nonetheless, the approach can be extended to include different severities in a straightforward way. An indicator for risk is then the probability function $P_{\text{coll}}(s;t,\Delta t)$ that a collision will happen during an interval of size $\Delta t$ around $t \hspace{0.0091cm} + \hspace{0.0091cm} s$. A compact risk measure $R(t)$ comprises, for each $t$, the entire accumulated expected future risk contained in $P_{\text{coll}}(s;t,\Delta t)$, during $s\in[0,\infty)$.  

\begin{figure*}[t!]
  \vspace{0.37cm}
  \centering
  \resizebox{0.75\linewidth}{!}{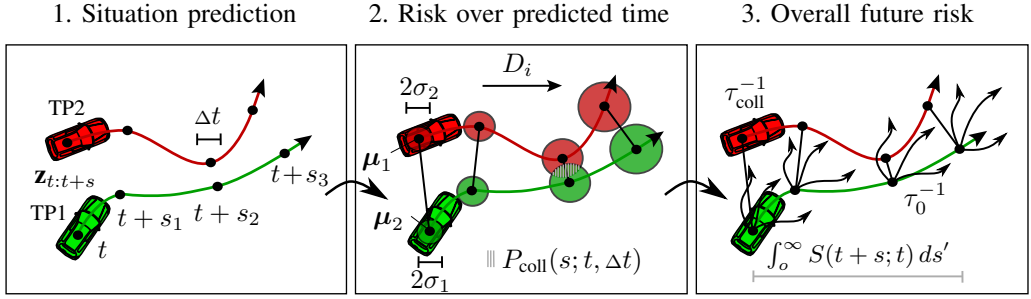}
  \caption{Combination of Gaussian position uncertainties and survival analysis for risk prediction.}
  \label{fig:combgs}
\end{figure*}

The RSD framework consists of three components as pictured in Figure \ref{fig:combgs}. In a first step, a prediction of how the situation will evolve in the future is calculated. In our notation, designating $\textbf{z}$ as the state vector of a  scene, the predicted sequence of future scene states is given by $\textbf{z}_{t:t+s}$. 
The prediction is thereby modeled with the help of road geometry information to constrain the paths on which vehicles can drive
and by a longitudinal velocity model in kinematic equations, with constant velocity as the easiest model. 
In a second step, $\textbf{z}_{t:t+s}$ is evaluated in terms of criticality.  
For this purpose, the normalized probability densities for the respective spatial positions of two TP's indexed $i=1,2$ are 
described by Gaussian functions 

\begin{equation}\label{mygauss}
f_{i}(x)=\frac{1}{\sqrt{2\pi\sigma^2_{i}}} \, \exp\left\{-\frac{(x-\mu_{i})^2}{2\sigma_{i}^2}\right\}
\end{equation} 

\noindent with the mean positions $\mu_{i}$ and variances $\sigma^2_{i}$.\footnote{The univariate bell curve is described, but generalizations to the bivariate case are analog.}
A collision at a position $x$ then occurs if both TP's coincide at the same position. Consequently, a way to quantify the likelihood 
of a collision at a common position $x$ is $f_{\text{coll}}(x)=f_1(x) f_2(x)$.
The probability that the first TP, driving along its trajectory, is hit by the second TP is eventually given by spatially 
integrating $f_{\text{coll}}(x)$ over all positions where the first TP can be 
\begin{align} \label{eq:prodgauss1d}
P_{\text{coll}}(s;t,\Delta t) &\sim \int_{\infty} f_{\text{coll}}(x) \, dx  \nonumber \\ 
= \frac{1}{\sqrt{2\pi(\sigma^2_{1}+\sigma^2_{2})}}\, & \exp\left\{-\frac{(\mu_2-\mu_1)^2}{2(\sigma^2_{1}+\sigma^2_{2})}\right\}.
\end{align}

The moving TP's follow a trajectory which undergoes certain variations in speed and 
heading. This accounts for mean positions through time $\mu_{i}(t+s)$ and growing spatial variances $\sigma^2_{i}(t+s)$. For
$\sigma^2_{i}(t+s)$, a simple Brownian motion diffusion model with constants $\sigma^2_{0,i}$ and $D_{i}$ is used
\begin{equation} \label{eq:diffsigma}
\sigma^2_{i}(t+s):=\sigma^2_{0,i} + D_{i} s.
\end{equation}

In the last step, accident occurrences are modeled as a thresholding process based on a Poisson-like event probability. 
Our inhomogeneous Poisson process is defined by a state-dependent total event rate $\tau^{-1}(\textbf{z}_{t:t+s})$, which characterizes the mean time between events and consists of a critical event rate $\tau^{-1}_{\text{crit}}$ and an escape rate $\tau^{-1}_0$ (any type of influences or behavioral options that contribute to mitigate resp.~"escape" from critical events) 
\begin{align}
\tau^{-1}(\textbf{z}_{t:t+s})=\tau^{-1}_0  &+ \tau^{-1}_{\text{crit}}.
\label{eq:escrate}
\end{align}

\noindent Here, we consider collision risk to be represented by the event rate $\tau_{\text{coll}}^{-1}$ for one TP pair. However, situations in which the ego car interacts with several other TP's $j$  and other types of risks, such as the risk of losing control in curves $\tau^{-1}_{\text{curv}}$, may be included likewise
\vspace{-0.1cm} 
\begin{align}
\tau^{-1}_{\text{crit}} = \sum_j  \tau^{-1}_{\text{coll,j}}+\tau^{-1}_{\text{curv}},
\end{align}
\vspace{-0.2cm}
\begin{equation}
\tau_{\text{coll}}^{-1}(\textbf{z}_{t:t+s}) = P_{\text{coll}}(s;t,\Delta t) / \Delta t.
\end{equation}

A so-called "survival function" indicates the probability that the vehicle will not be engaged in an event like an accident 
from $t$ until $t+s$ and is given by
\begin{equation}
S(s;t,\textbf{z}_{t:t+s})=\exp\{-\int_o^s \tau^{-1}(\textbf{z}_{t:t+s'}) \, ds'\}.
\label{eq:survfunc}
\end{equation}

\noindent Combining Eq. (\ref{eq:survfunc}) with Eq. (\ref{eq:escrate}) as in \cite{eggert2014}, one can derive a probability density for general events 
\begin{equation}
p_{E}(s;t,\textbf{z}_{t:t+s}) = \tau^{-1}_0S(s;t,\textbf{z}_{t:t+s})+ \tau^{-1}_{\text{crit}}S(s;t,\textbf{z}_{t:t+s})
\end{equation}
and obtain the overall future risk as the integral over all predicted times of the critical events only 
\begin{equation}\label{IntRisk}
R(t) = \int_0^{\infty} \tau_{\text{crit}}^{-1}(\textbf{z}_{t:t+s})S(s;t,\textbf{z}_{t:t+s})\,ds.
\end{equation} 
Due to numerical reasons, the actual temporal integration is capped with a fixed prediction horizon $s_{\text{max}}$.

\subsection{Uncertainties in Naturalistic Driving} 
\label{secsec:extensions}

In RSD, the Gaussian probability densities allow to formulate collision uncertainty as a function of spatial uncertainties $\sigma^2_{i}(t+s)$ from the involved TP's. The value of $\sigma^2_{0,i}$ quantifies measurement uncertainty and can be different for each TP. It reflects the uncertainty at the current time with $s\hspace{-0.03cm}=\hspace{-0.03cm}0$. For future times $s>0$, the parameter $D_i$ specifies prediction uncertainty. Subsequently, the survival analysis normalizes the event probabilities related to all TP's and risk types and for the time $s_E$ until a critical event 
\begin{equation}
\lim_{s_E \rightarrow 0} \,R(t)\rightarrow 1 \ \text{and} \lim_{s_E \rightarrow \infty} \,R(t)\rightarrow 0
\end{equation}
holds true. A constant escape rate $\tau^{-1}_0$ reduces thereby $S(s;t,\textbf{z}_{t:t+s})$ over the predicted time and introduces an accumulating event avoidance effect (events in the more distant future are considered to a lesser extent). Similarly, if high $\tau^{-1}_{\text{crit}}$ occurs early, for all times afterwards $S(s;t,\textbf{z}_{t:t+s})$ is diminished. RSD thus takes historical uncertainty correctly into account (future risks that arise after another critical event are further reduced).  

As shown in \cite{eggert2017}, longitudinal following and intersection crash cases are detected by RSD earlier than by the Gaussian method alone or TTC (at least $\unit[1.1]{sec}$ before the critical event actually happens).\footnote{As comparison, the Gaussian method had a minimal detection time of $\unit[0.8]{sec}$ and TTC could identify only longitudinal crashes $>\unit[0.7]{sec}$ ahead.} At the same time it has a considerably lower number of false positive detections for near- and non-crash cases.\footnote{In the experiments, we used an event threshold of $R(t)\hspace{-0.05cm}>\hspace{-0.05cm}0.7$ for up-coming accidents. Decreasing the threshold would lead to higher sensitivity of RSD for accidents.} Nevertheless, to better quantify the predicted risks in normal TP driving, RSD needs to precisely account for further uncertainties. For this purpose, in the following Subsections \ref{secsecsec:2Dgaussians}, \ref{secsecsec:veluncertainty} and \ref{secsecsec:mixturemodel}, 
we introduce extensions of the uncertainty models that improve RSD in three special situations: close passing of other TP's in 1. longitudinal segments, 2. when stopping in front of as well as 3. turning at intersections.

\begin{figure}[t!]
  \begin{center}
    
    \resizebox{0.7\linewidth}{!}{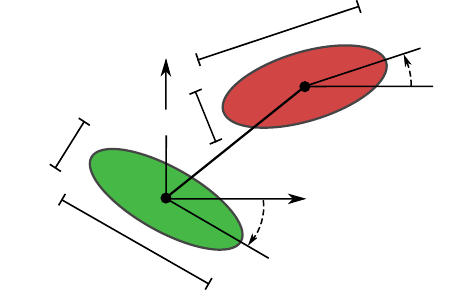} 
    \vspace{1.35cm}
    
    \resizebox{0.65\linewidth}{!}{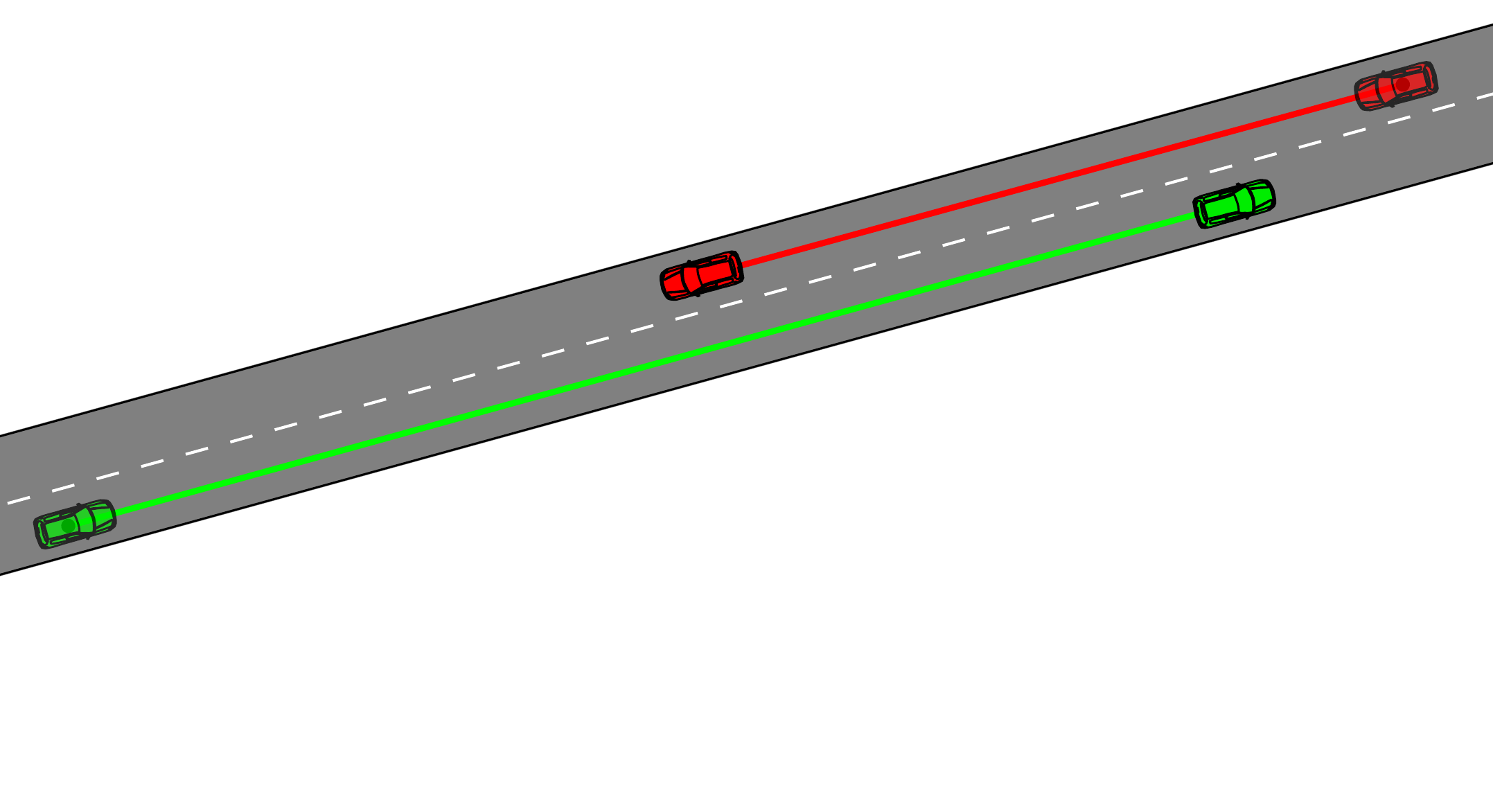}
    \vspace{-0.5cm}
    \caption{Top: Schema and variables of Gaussian ellipses. Bottom: Unharmful passing another TP on straight road.}
    \label{fig:2dgauss}
  \end{center}
\end{figure} 

\vspace{0.25cm}
\subsubsection{2D Gaussians} 
\label{secsecsec:2Dgaussians}

Because the TP's are predicted to drive along predefined paths, we assume that $\sigma^2_{i}(t+s)$ has to be extended to 2 dimensions for handling longitudinal and lateral influences. 
In this way, we obtain ellipses  
specified by an uncertainty matrix $\mathbf{\Sigma}^\text{'}_{i}$ around the mean 
position vector $\boldsymbol{\mu}_{i}$. 
\begin{equation}
\boldsymbol{\mu}_{i} = 
  \begin{bmatrix}
    {\mu}_{x,i} \\
    {\mu}_{y,i} \\
  \end{bmatrix}, \quad
\mathbf{\Sigma}^\text{'}_{i} = 
  \begin{bmatrix}
    \sigma^2_{\text{lon},i} & 0 \\
    0 & \sigma^2_{\text{lat},i} \\
  \end{bmatrix} 
\label{eq:musigma}
\end{equation}

The top of Figure \ref{fig:2dgauss} shows the 2D Gaussians for one point in time $t\hspace{-0.04cm}+\hspace{-0.04cm}s$. The relative orientations to the absolute $x,y$-coordinate system are indicated with $\alpha_{i}$.
To retrieve the product of the corresponding Gaussian functions $\mathbf{f}_i$, the longitudinal and lateral uncertainties $\mathbf{\Sigma}^\text{'}_{i}$ have to be transformed with
\begin{equation}
\mathbf{\Sigma}_{i} = \mathbf{R} \mathbf{\Sigma}^\text{'}_{i} \mathbf{R}^T \ \text{and} \
\mathbf{R} = 
  \begin{bmatrix}
    \cos \alpha_{i} & -\sin \alpha_{i} \\
    \sin \alpha_{i} & \cos \alpha_{i} \\
  \end{bmatrix}. 
\label{eq:rotsigma}
\end{equation}
\noindent Equation (\ref{eq:prodgauss1d}) can then be rewritten in 2D to 
\begin{align}
P_{\text{coll}}(s;t,\Delta t) = |2\pi&(\mathbf{\Sigma}_{1}+\mathbf{\Sigma}_{2})|^{-\frac{1}{2}} * \nonumber \\ \exp\{-\frac{1}{2}
(\boldsymbol{\mu}_{2}-\boldsymbol{\mu}_{1})^T(\mathbf{\Sigma}_{1}&+\mathbf{\Sigma}_{2})^{-1}
(\boldsymbol{\mu}_{2}-\boldsymbol{\mu}_{1})\}.   
\label{eq:prodgauss}
\end{align}

Without incorporating the orientation of the TP's, a longitudinal scenario of two TP's passing closely with a lateral constant offset from time $t_{\text{start}}$ to $t_{\text{end}}$ (see bottom of Figure \ref{fig:2dgauss}) has the same high risk as an intersection scenario of two TP's passing closely with 90$^{\circ}$. By contrast, elongated 2D Gaussians rate longitudinal passing as safe. 

\vspace{0.25cm}
\subsubsection{Position Uncertainty by Velocity Variance} 
\label{secsecsec:veluncertainty}

\noindent Over the predicted time, $\sigma_{i}(t+s)$ grows proportionally to $\sqrt{s}$ according to Equation (\ref{eq:diffsigma}). 
We extrapolate the kinematics of the current state to retrieve trajectories, but the velocities of the TP's
are not influencing the uncertainty prediction. After a prediction step of size $\Delta s$, their longitudinal position on the path $l_{i}$
is shifted by $\Delta l_{i}$ according to 
\begin{equation}
l_{i}(s+\Delta s) = l_{i}(s) + \Delta l_{i} = l_{i}(s) + v_{i}(s) \Delta s
\label{eq:diststep}
\end{equation}
with velocities $v_i$.\footnote{Remark: we set the current time $t=\unit[0]{sec}$ and look only at the increment in the predicted time $s$.}
When we additionally assume a longitudinal velocity uncertainty according to e.g. a normal distribution with variance $\sigma_{v,i}=\langle v_i \rangle c_i$, the increase of spatial uncertainty is determined by the velocity uncertainty factor $c_i$. For a discrete step in prediction time we then get   
\begin{align}
\sigma_{l,i}(s+\Delta s) := \sigma_{l,i}(s) &+ c_{i} v_{i}(s) \Delta s, 
\label{eq:sigmavstep}
\ \\[5pt] 
\vspace{1.5cm}
\sigma_{l,i}(s=0) &= \sigma_{0,i}.
\end{align}
Here, 
$\sigma_{l,i}(s)$ becomes essentially proportional to $s$.  

\begin{figure}[t!]	
  \begin{center}
    \resizebox{0.7\linewidth}{!}{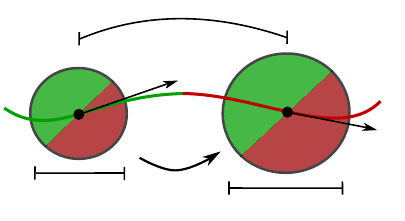} \\
    \vspace{-0.35cm}
    \resizebox{0.65\linewidth}{!}{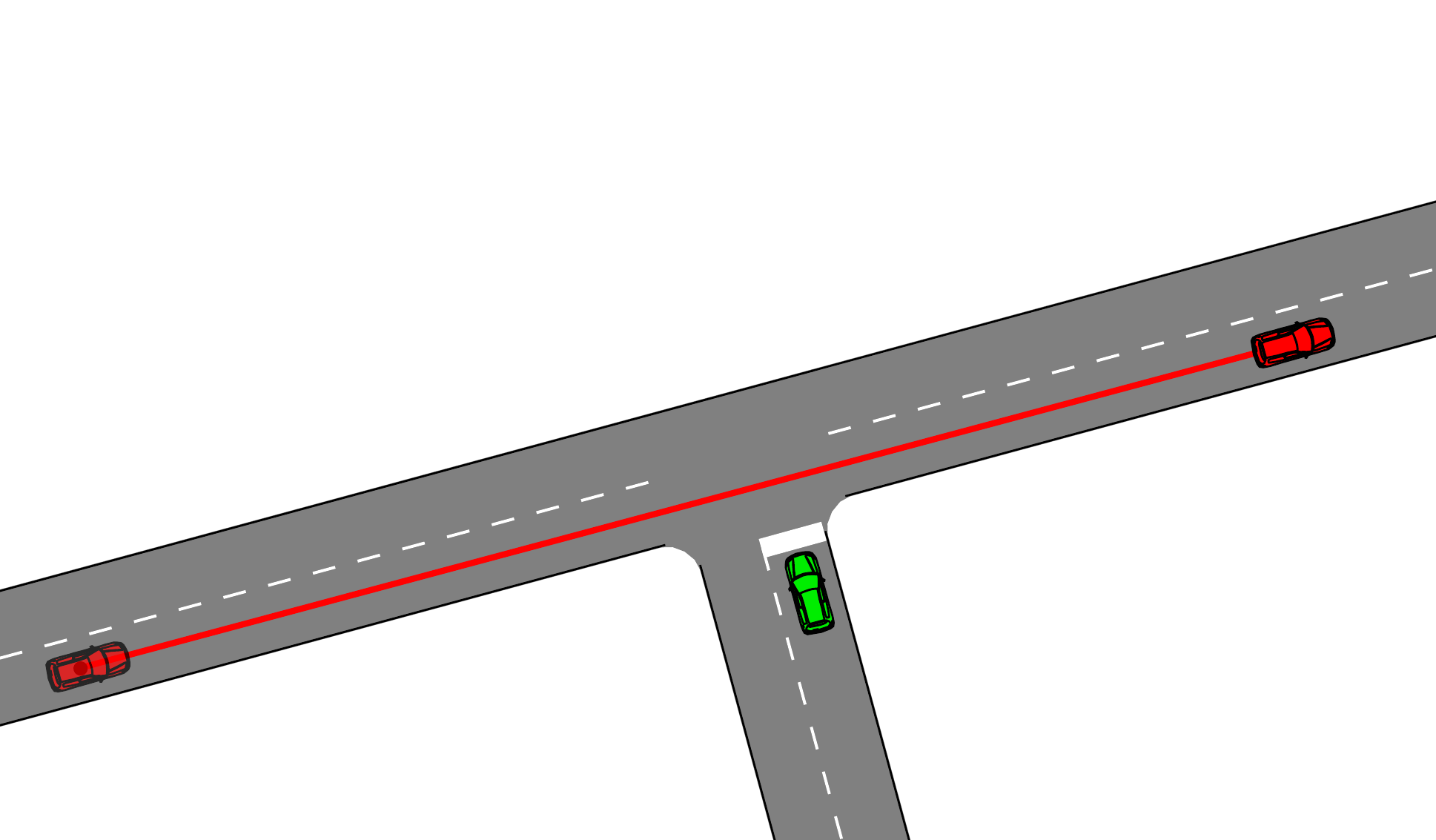}
    \caption{Top: Velocity-dependent Gaussian growth over prediction time. Bottom: Securely stopping at intersection, while another TP crosses.}
    \label{fig:vel_uncertainty}
  \end{center}
\end{figure}

Especially in scenarios with extreme velocities (i.e., $v_i\hspace{-0.07cm}<\unit[5]{m/sec}$ and $v_i\hspace{-0.07cm}>\hspace{-0.07cm}\unit[15]{m/sec}$), the previous Brownian diffusion model lead to over- or underestimation of uncertainties and thus the contained risk.
Figure \ref{fig:vel_uncertainty} outlines the change in the width of the Gaussians $2\sigma_{l,i}(s)$ and a waiting TP at a T-intersection while another TP is crossing during the time interval $[t_{\text{start}}, t_{\text{end}}]$. Although the situation can be categorized as safe, Brownian position uncertainty would lead to large risk areas around the standing green car position. To the contrary, the velocity propagation approach yields constant small $\sigma_{l,i}(s)$ for the stopped car and classifies this situation as non-critical.

\begin{figure}[t!]
  \begin{center}
    \resizebox{0.7\linewidth}{!}{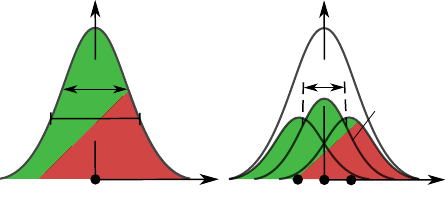} \\

    \vspace{0.2cm}
    \resizebox{0.65\linewidth}{!}{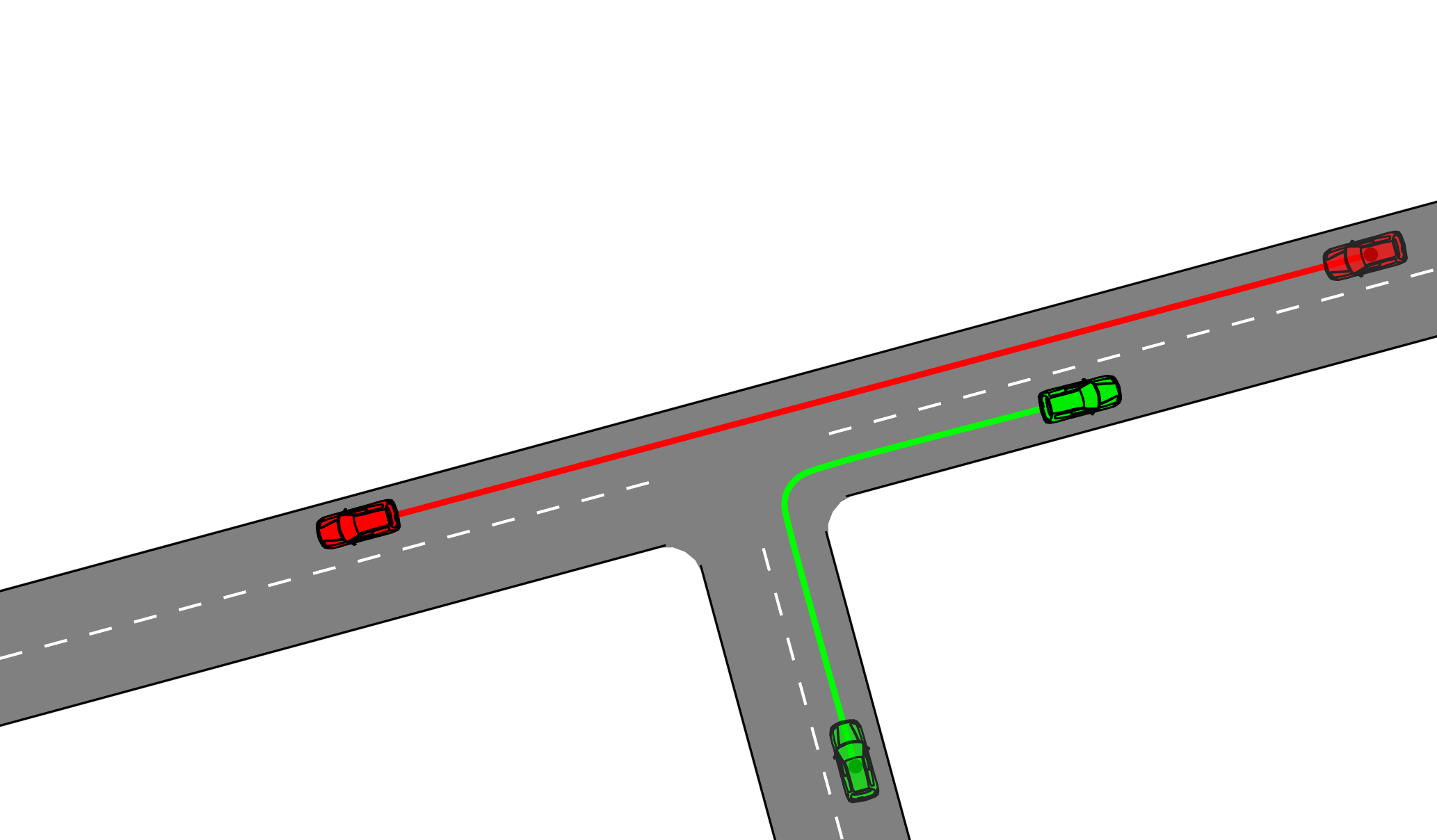}
    \caption{Top: Partition of Gaussian into multiple components. Bottom: Safely turning at intersection with oncoming TP.}
    \label{fig:gmm}
  \end{center}
\end{figure} 

\vspace{0.25cm}
\subsubsection{Path-following Mixture Model}
\label{secsecsec:mixturemodel}

\noindent Large $\sigma_{i}(t+s)$ at prediction times $s\gg0$ might unintentionally cover opposite lanes 
when a TP is turning. Therefore, we include geometry-sensitive Path-following Mixture Models (PMM) allowing curved Gaussian shapes for position uncertainties \cite{lin1999}. 

First, we split the Gaussian with $2\sigma_{i}$ into $N$ smaller Gaussians of $2\sigma_{i,k}$ as drawn in the upper half of Figure \ref{fig:gmm}. The smaller Gaussian components at $\mu_{i,k}$ are spread to the right and left from $\mu_{i}$, while intersecting close to their Full Width at Half Maximum $\text{\fontsize{8}{12}\selectfont FWHM}$ value.\footnote{At the same time, we require one element with $\mu_{i,k}=\mu_{i}$ and thus an uneven number $N$.} The composition heuristics are summarized by
\begin{equation}
\sigma_{i,k} = \frac{m_f \text{\fontsize{8}{12}\selectfont FWHM}}{N} \sigma_{i},
\label{eq:sigma_gmm}
\end{equation}
\vspace{-0.5cm}
\begin{align}
&\mu_{i,k} = \mu_{i} + \frac{2k}{m_f\text{\fontsize{8}{12}\selectfont FWHM}} \sigma_{i,k} \\
\ \text{with} \ k=0&,...,\pm\frac{N-1}{2} \ \text{and} \ \text{\fontsize{8}{12}\selectfont  FWHM}=2\sqrt{2\ln{2}}. \nonumber 
\label{eq:mu_gmm}
\end{align}
The constant $m_f$ ensures that the constructed PMM does not contain additional local minima and that it has a smooth shape.\footnote{For $m_f>1$, the mixture components get closer to each other and have higher $\sigma_{i,k}$.} 

At last, each function of the component $f_{i,k}$ is summed up and weighted with the factor $w_{i,k}$ to reduce the deviation from the former Gaussian $f_{i}$, which leads to

\begin{equation}
w_{i,k} = f_{i}(\mu_{i,k}) \frac{f_{i}(\mu_{i})}{\sum_{k} f_{i}(\mu_{i,k}) f_{i,k}(\mu_{i})},
\label{eq:weights_gmm}
\end{equation}
\begin{equation}
f_{\text{pmm},i}(x) = \sum_{k} w_{i,k}f_{i,k}(x).
\label{eq:f_gmm}
\end{equation}

\noindent In $w_{i,k}$, the collective peaks of the components $\sum_{k} f_{i,k}(\mu_{i,k})$ are scaled to match the desired original heights $f_{i}(\mu_{i,k})$. 
Our PMM composition heuristics achieves similar reconstruction errors as parameter optimization approaches for $f_{\text{pmm},i}(x)$ and $N>13$, however providing a fast and direct calculation. 

In the lower half of Figure \ref{fig:gmm}, one TP takes a sharp curve and another TP crosses the intersection. During the situation occuring from $t_{\text{start}}$ until $t_{\text{end}}$, the resulting phantom risks from the elongated Gaussians are avoided with the PMM. 

\section{Behavior Extrapolation}
\label{sec:behavioruncertainty}
For longitudinal collisions, Time Headway (TH) or Time-To-Collision (TTC) are able to quantify risks in terms of (inverse) time until a critical event happens. In this Section, we describe their underlying formulas and describe how the properties of RSD can be used to cover risks in more general terms and in particular to integrate the prediction assumptions from both TH and TTC.

\subsection{TH and TTC}
While an ego vehicle is driving with longitudinal velocity $v_1$ along a path and following another vehicle, TH \cite{transpres2010} describes the time until the ego car travels from the current longitudinal position $l_1$ to the current longitudinal position of the other car $l_2$ according to\footnote{Here, index $i=1$ always denotes the ego car and $i=2$ the next car in front.}
\begin{equation}
\text{TH} = \frac{-\Delta l}{v_1} \ \text{with} \ \Delta l = l_1 - l_2. 
\label{eq:th}
\end{equation}

\noindent The average human reaction time lies around $t_r \approx \unit[1]{sec}$ \cite{winner2015}. By keeping $\text{TH} > t_r$, once the other entity brakes at $l_2$, the ego entity has some time left to apply an appropriate deceleration to mitigate or even avoid a collision. 


TH can be seen as a risk measure. Its inverse $1/\text{TH}$ is assumed to correlate with the collision probability given that the follower drives with constant $v_1$ in combination with the front vehicle suddenly stopping at $l_2$. By contrast, TTC \cite{horst91} assumes that both cars continue driving with constant longitudinal velocities $v_1$ and $v_2$ and calculates the time until a collision occurs when $l_1=l_2$,

\begin{equation}
\text{TTC} = \frac{-\Delta l}{\Delta v}, \text{whereby} \ \Delta v = v_1 - v_2.
\label{eq:ttc}
\end{equation}

\noindent The left and middle part of Figure \ref{fig:thttcrsd} visualize the longitudinal car following scenario and the designated collision points in TH and TTC.
For forward driving, a valid TTC only exists for $-\Delta l>0$ and $\Delta v >0$. In this case, $v_1 >\Delta v$ holds so that TH always overestimates potential risks as compared to TTC (i.e., $1/\text{TH}>1/\text{TTC}$). 

\subsection{Behavior Uncertainty}
\label{secsec:behaviorunc}

TH and TTC approximate the critical event probabilities based on the point of maximal criticality by deriving one collision time with simple kinematic equations. If the hypothetical accident does not occur as for non-longitudinal scenarios, there is no risk at all. Some attempts have been made to make TTC more flexible with e.g. extrapolating constant deceleration for the obstacle \cite{winner2015}. Analogously, the required longitudinal acceleration of the follower can be compared with its maximal possible value to calculate Brake Threat Numbers (BTN) \cite{brannstrom2008}. 

Nevertheless, RSD has the beneficial property of estimating a continuous, differential, probabilistic risk along the entire future predicted time, see Eq.~(\ref{IntRisk}). Furthermore, it is valid independently of the predicted behavior and thus can be used with the same prediction assumptions from TH or TTC, but also for completely arbitrary paths and velocity profiles.
With RSD, the velocity-dependent position uncertainty $\sigma_{l,i}(s)$ allows to systematically incorporate behavior uncertainty. For constant velocity assumptions, this results in the front part of the probability densities being fed by acceleration and the back part by deceleration behaviors. The middle of the probability densities around $\boldsymbol{\mu}_{i}$ constitutes constant average velocity. 

\begin{figure}[t!]
  \begin{center}
    \vspace{0.7cm}
    \resizebox{\linewidth}{!}{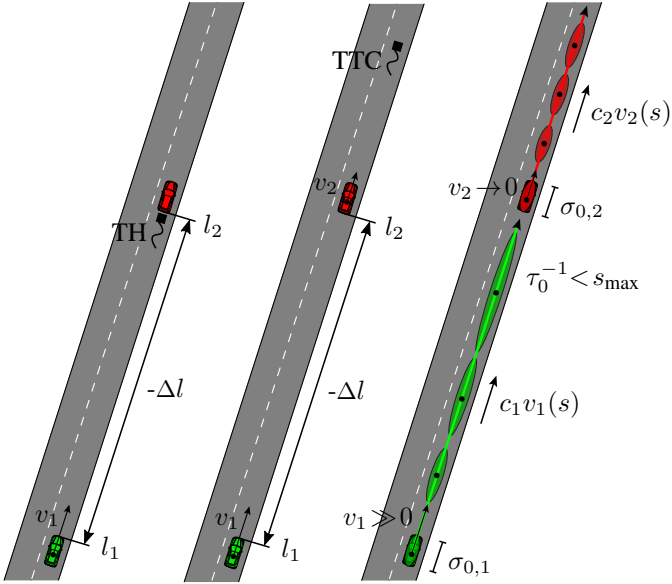} \\
    \caption{Left: Kinematic variables for TH. Middle: Additional velocity for the other car with TTC. Right: Parameter setting of behavior uncertainty in RSD.} 
    \label{fig:thttcrsd}
  \end{center}
\end{figure}

We set $s_{\text{max}}= \unit[12]{sec}$ and $\tau^{-1}_0= \unit[3]{sec}$ to be capable of incorporating other TP's at least $\text{TH}<\unit[5]{sec}$ far away. Then, the motion planner Risk Optimization Method (ROPT) \cite{puphal2018} was employed to parametrize $\sigma_{0,i}$ and $c_i$. ROPT utilizes RSD in a cost function to find safe behaviors through traffic. For longitudinal and intersection scenarios, we ran sanity checks whether ROPT holds reasonable distance thresholds to other TP's in various $v_{i}$ settings. Increasing $c_i$ lets ROPT keep longer distances. In addition, we looked at typical accelerations at $s=0$ plus velocity changes after $s=\unit[3]{sec}$ in NGSIM and fitted their standard deviations to $c_i$. In both procedures, we obtained $6\sigma_{0,i}\hspace{-0.03cm}=\hspace{-0.03cm}\unit[4]{m}$ as the average TP length\footnote{Equations (\ref{eq:th}) and (\ref{eq:ttc}) are both extended to consider the car sizes by changing $\Delta l$ to $\Delta l^* = \Delta l + \unit[4]{m}$.}
and $c_i\hspace{-0.03cm}=\hspace{-0.03cm}0.1$.

The length increase of the Gaussians over the predicted time differ according to $v_i(s)$. In the example of Figure \ref{fig:thttcrsd} (see right part), the velocity of the follower $v_1$ is larger than $v_2$ of the front TP. In other words, the ellipses grow for TP1 while staying nearly constant for TP2. 
We found that the fitted parameter setting covers the average behavior uncertainty of normal NGSIM driving statistics well, resulting in a mixture of the sudden stop assumption from TH and the constant velocity prediction from TTC. 

In the RSD formalism, positional uncertainties between TP's are not explicitly correlated. Aside from this, the constant velocity prediction suggests unawareness of the TP's from each other. However, the risk calculation incorporates a mutual influence in TP's by the spread of the Gaussian velocity distributions and thus assuming that vehicles can take on velocities that deviate from constant velocity (e.g. a follower that accelerates onto an obstacle in front that brakes and vice versa). Furthermore, since RSD is agnostic to the type of predicted trajectories, additional interactions can be incorporated by modeling particular motion patterns.

\section{Simulations}
\label{sec:experiments}
\subsection{Lankershim Boulevard}
\label{secsec:lankershim}

The NGSIM dataset \cite{ngsim2006} consists of 5 traffic study areas in the US, in which cameras are mounted on high buildings. With computer vision techniques, the positions of all vehicles were detected with an accuracy of $(\Delta x\hspace{-0.11cm}=\hspace{-0.11cm}\unit[0.6]{m}, \Delta y \hspace{-0.11cm}=\hspace{-0.11cm} \unit[1.2]{m})$. 
In particular, Lankershim Boulevard is suitable for testing the robustness of RSD. It has multiple intersections (4 with traffic lights, 2 with priority), wide and narrow road structures (2, 3 and 4 lanes), inner-city and highway stretches (velocities in the range $v \hspace{-0.1cm}= \hspace{-0.1cm}\unit[0\text{-}20]{m/sec}$), a long curve (making up $20 \%$ of the total section) and dense traffic (1000 vehicles in 15 minutes). As $96 \%$ of the TP's are cars, it is reasonable to neglect the particular masses and sizes ($3.5 \%$ trucks plus buses and $0.2 \%$ motor bikes). 

For the risk analysis, we first loaded and normalized the trajectories \cite{saunier2018}. By employing an exponential moving average filter forwards and backwards
, we afterwards separately smoothed the positions $p$, velocities $v$ and accelerations $a$. Both $v$ and $a$ are gained by differentiation from $p$, which results in additional noise. Consequently, we set different smoothing widths $T_p\hspace{-0.06cm}=\hspace{-0.06cm}\unit[10]{sec}$, $T_v\hspace{-0.06cm}=\hspace{-0.06cm}\unit[20]{sec}$ and $T_a\hspace{-0.06cm}=\hspace{-0.06cm}\unit[80]{sec}$ to compensate the effect \cite{thiemann2008}. As a next step, we successively assumed each car to take the role of an ego car and extracted all other cars in the same time interval. 
In the simulation, we eventually took the real driven position sequence as paths for the trajectory prediction of RSD, TH plus TTC. 
In this way, we already know in advance the intentions of the cars (e.g. lateral lane changes or turn at intersection). Because of fixed velocity extrapolations (i.e., constant velocity or sudden stop), their longitudinal behavior is however assumed to be unknown. Risks can only come from wrongly applied velocities of the cars along their paths.

\begin{figure}[t!]
  \centering
  \resizebox{0.6\linewidth}{!}{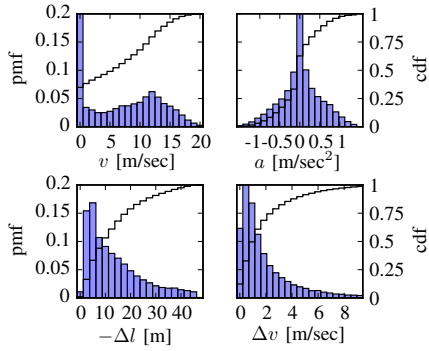}
  \vspace{0.12cm}
   \caption{Variance of kinematics in data extracted from NGSIM (Lankershim Boulevard).}
    \label{fig:ngsim_statistics}
\end{figure}

Figure \ref{fig:ngsim_statistics} shows the resulting probability mass function $\text{pmf}$ and cumulative distribution function $\text{cdf}$ with a histogram representation for the occurring $v$ and $a$ as well as distances $-\Delta l$ and relative velocities $\Delta v$ to the front vehicle.\footnote{Looking exemplarily at the histogram for $v$, pmf represents the share of data points within an interval of $\unit[1]{m/sec}$ and cdf is the successive, accumulative sum of pmf values from left to right.} 
More than $25\%$ of the time, the vehicles stand in traffic or in front of an intersection. The velocity distribution is bimodal, with a high peak at $v\approx\unit[0]{m/sec}$ and a broader peak around $v=\unit[12]{m/sec}$. Overall, vehicles do not excessively brake or accelerate. The acceleration distribution has its mean at $\mu_a=\unit[0]{m/sec^2}$ with a deviation of $\sigma_a=\unit[0.4]{m/sec^2}$. When there is a front vehicle, most $-\Delta l$ lie around $\mu_{\Delta l}=\unit[5]{m}$ of a possible log-normal distribution. Only $5\%$ have lower values than $\mu_{\Delta l}$ and these happen at small $v$. In contrast, $\Delta v$ correspond to a logarithmic distribution with $\mu_{\Delta v}=\unit[1]{m/sec}$, whereby big $\Delta v$ take place in any interval of $v$. 

\begin{figure}[t!]
  \centering
  \resizebox{1.0\linewidth}{!}{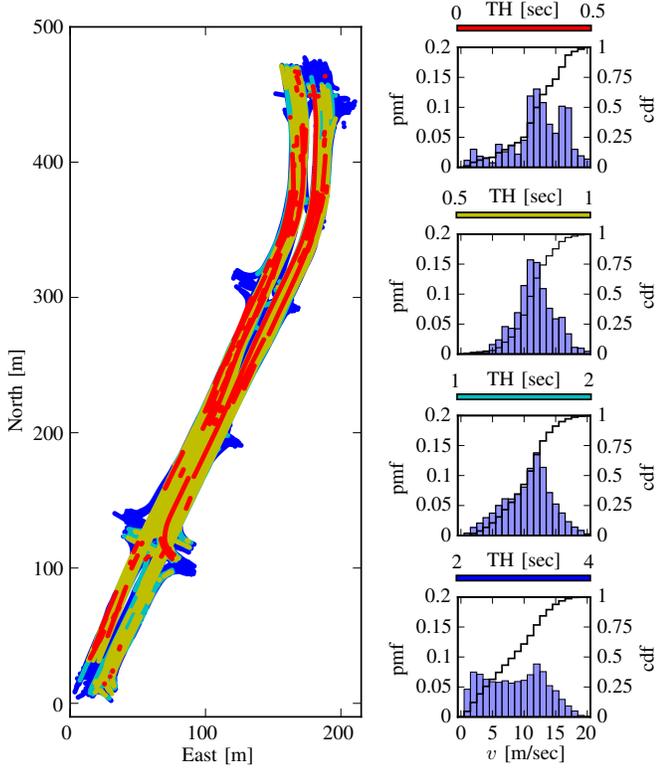}
  \caption{Criticality map and velocity histograms of TH. One can see that TH detects tailgating on highways.} 
  \label{fig:riskmap_th}
\end{figure} 

\subsection{Results}
In the following, we analyze the qualititave and quantitative differences between the risk measures based on RSD, TH and TTC. The experimental goal is to reason about which driving behaviors may create potential fatal outcomes and to find the corresponding street areas on which they likely appear. For this purpose, we define four criticality bins (dangerous, offensive, uncomfortable, noticeable) and set them for TH accordingly to $b_1\hspace{-0.05cm}=\hspace{-0.05cm}[\unit[0]{sec}, \unit[0.5]{sec}]$, $b_2\hspace{-0.05cm}=\hspace{-0.05cm}[\unit[0.5]{sec}, \unit[1]{sec}]$, $b_3\hspace{-0.05cm}=\hspace{-0.05cm}[\unit[1]{sec}, \unit[2]{sec}]$ and $b_4\hspace{-0.05cm}=\hspace{-0.05cm}[\unit[2]{sec}, \unit[4]{sec}]$.\footnote{As references, in ACC the minimal TH is $\unit[1]{sec}$ and the maximal $\unit[3]{sec}$. Furthermore, $\text{TH}\hspace{-0.05cm}=\hspace{-0.05cm}\unit[2]{sec}$ is recommended on US highways.} 
The bins are colored red, yellow, cyan and blue and every calculated TH value of the NGSIM trajectories with discretization of $\Delta t \hspace{-0.05cm}=\hspace{-0.05cm}\unit[0.1]{s}$ is sorted into them.\footnote{Approximately $40\%$ of the data points are in the bins, the rest $60\%$ reflect safe behaviors with no noticeable risk at all.} For RSD and TTC, we then fill the bins consecutively with the most to least hazardous events until they contain the same amount of events as for TH. The boundaries of the bins arise automatically from the corresponding first and last data points sorted into each bin. 
Eventually, we create criticality maps plotting the most riskful color of RSD, TH and TTC at each road point and analyze the velocity distributions in their bins. 

\begin{figure}[t!]
  \centering
  \resizebox{1.0\linewidth}{!}{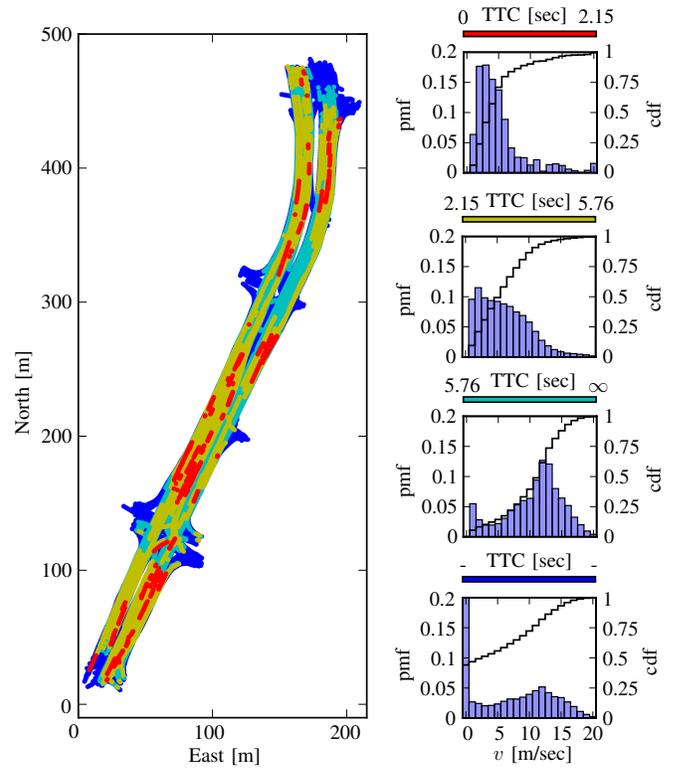}
  \caption{Criticality map and velocity histograms of TTC. TTC captures well braking risks at intersections.} 
  \label{fig:riskmap_ttc}
\end{figure} 

The outcome for TH is depicted in Figure \ref{fig:riskmap_th}. TH classifies dangerous and offensive risks at velocity intervals around a Gaussian with $\mu_{\text{TH}}=\unit[12]{m/sec}$ and $\sigma_{\text{TH}}=\unit[2.5]{m/sec}$. When the ego car is following another car with high $v$ and relatively low $|\Delta l|$, the sudden stop prediction causes strong worst cases. These situations are located mostly on the top right curvy segment. Moreover, uncomfortable and noticeable TH appear also for interplays of an ego car braking from moderate $v$ approaching another car waiting close to an intersection. Due to the discontinuity of Equation (\ref{eq:th}), TH cannot evaluate risks for $v\hspace{-0.05cm}\rightarrow \hspace{-0.05cm}0$. 

\begin{figure}[t!]
  \centering
  \resizebox{1.0\linewidth}{!}{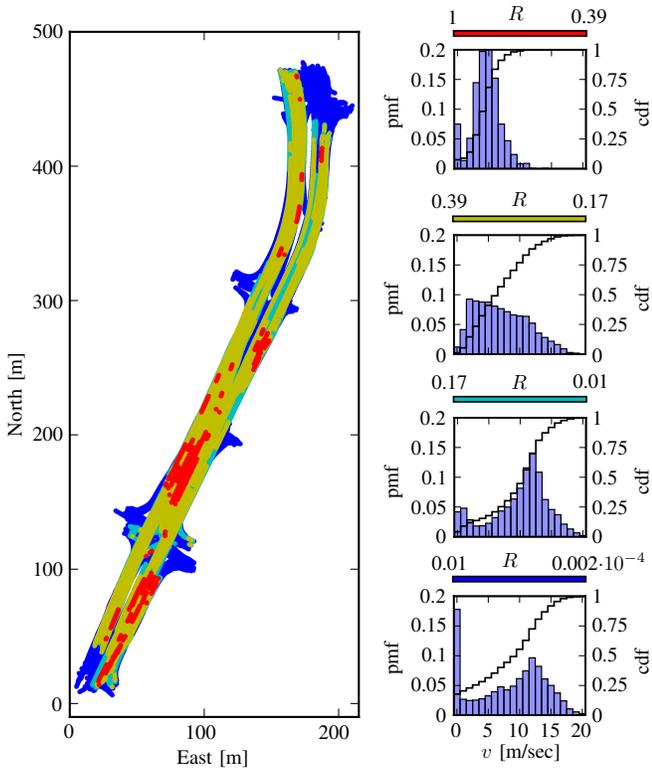}
  \caption[Criticality map and velocity histograms of RSD for front car. RSD identifies intersection braking and highway tailgating.]{Criticality map and velocity histograms of RSD for front car. RSD identifies intersection braking and highway tailgating.\footnotemark} 
  \label{fig:riskmap_rsd_front}
\end{figure} 

\footnotetext{Note the difference in the boundaries of the last two bins to TTC in Figure \ref{fig:riskmap_ttc}. RSD is capable of having low but existent risk values, while TTC cannot distinguish all tailgaiting incidents from safe behaviors.}

Now we concentrate on TTC, which depends on $\Delta v$. 
Mainly for approaching the lowest large intersection ($\text{East}\hspace{-0.05cm}=\hspace{-0.05cm}\unit[70]{m},\text{North\hspace{-0.05cm}}=\hspace{-0.05cm}\unit[120]{m}$) with big $\Delta v$, TTC detects risks in the red and yellow bins (see Figure \ref{fig:riskmap_ttc}). Their boundaries lie at $b_{1,\text{TTC}}\hspace{-0.05cm}=\hspace{-0.05cm}[\unit[0]{sec}, \unit[2.15]{sec}]$ and $b_{2,\text{TTC}}\hspace{-0.05cm}=\hspace{-0.05cm}[\unit[2.15]{sec}, \unit[5.76]{sec}]$ in the common warning intervals of $\unit[1-5]{sec}$ for CMS. Alongside the logistic distribution of $\mu_{\text{TTC}}=\unit[3]{m/sec}$ and $\sigma_{\text{TTC}}=\unit[3]{m/sec}$, the cyan bin includes the critical car following scenario of TH. But since $b_{3,\text{TTC}}\hspace{-0.05cm}=\hspace{-0.05cm}[\unit[5.76]{sec}, \infty]$, the data points are
too few. TTC cannot extract car following incidents with small distances, since $\Delta v \approx 0$ in these cases. On that account, the last blue bin includes arbitrary $v$ in NGSIM. This can be seen by comparing the distribution shape of the last blue bin with the overall $\mbox{pmf}$ of $v$ from Figure \ref{fig:ngsim_statistics}.

Whereas TH and TTC only capture frontal longitudinal collision risks, RSD is able to capture all possible collision risks. However for a fair comparison, RSD is initally applied solely on the front TP. The same parametrization of $\sigma_{0,i}$ and $c_i$ from Section \ref{secsec:behaviorunc} is set for both the ego vehicle and other TP. As $R$ is normalized to $[0,1]$ and represents a probability, the first bin in Figure \ref{fig:riskmap_rsd_front} starts with $R=1$ and the last ends with $R\approx0$. The thresholds are as follows:
$b_{1,\text{RSD}}\hspace{-0.05cm}=\hspace{-0.05cm}[1, 0.39]$, $b_{2,\text{RSD}}\hspace{-0.08cm}=\hspace{-0.08cm}[0.39, 0.17]$, $b_{3,\text{RSD}}\hspace{-0.05cm}=\hspace{-0.05cm}[0.17, 0.01]$ and $b_{4,\text{RSD}}\hspace{-0.05cm}=\hspace{-0.05cm}[0.01, 0.002\cdot10^{-4}]$. 
To be compatible with TTC, we base the RSD on a constant velocity prediction. Also similarly to TTC, RSD sorts risk zones around intersections from existent $\Delta v$ in combination with decreasing $|\Delta l|$ at $\mu_{\text{RSD},1}\hspace{-0.03cm}=\hspace{-0.03cm}\unit[4]{m/sec}$ into $b_{1,\text{RSD}}$ and $b_{2,\text{RSD}}$. In addition, RSD incorporates acceleration and deceleration behavior from the velocity uncertainty. A distribution shape similar to TH appears with $\mu_{\text{RSD},2}=\unit[12]{m/sec}$ in $b_{3,\text{RSD}}$ and $b_{4,\text{RSD}}$. With large $c_i$, the abrupt stop prediction from TH can also be incorporated into RSD. This would shift the critical following incidents to $b_{1,\text{RSD}}$ and $b_{2,\text{RSD}}$ as well as the intersection approaching to $b_{3,\text{RSD}}$ and $b_{4,\text{RSD}}$. In general, constant velocity is however statistically more realistic (refer to $\mu_a=\unit[0]{m/sec^2}$ in Section \ref{secsec:lankershim}). 

\begin{figure}[t!]
  \centering
  \resizebox{1.0\linewidth}{!}{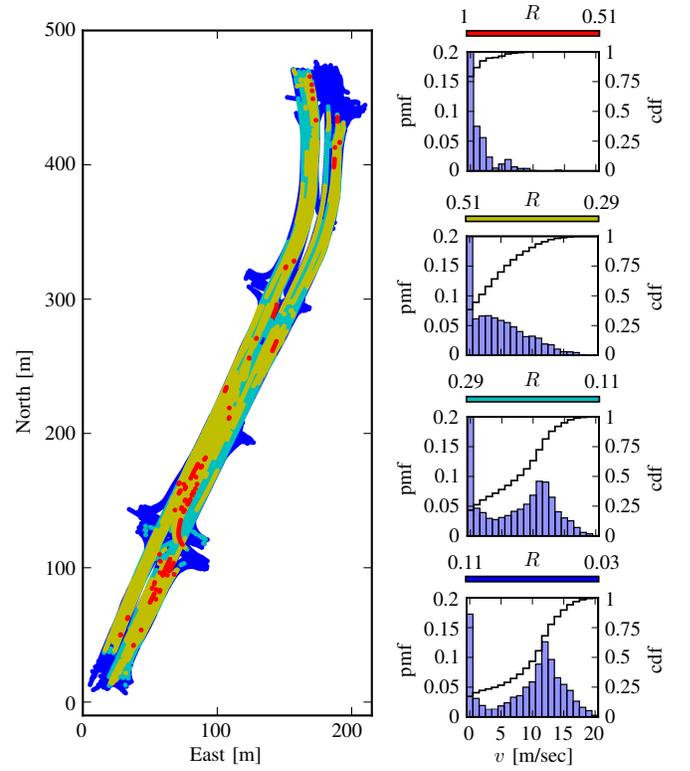}
  \caption{Criticality map and velocity histograms of RSD for all surrounding cars. Now, RSD has more localized hot spots and describes in each bin the relation to traffic density.} 
  \label{fig:riskmap_rsd_all}
\end{figure}

\vspace{0.25cm}
\subsubsection{Traffic Situation Risk}
We discovered that TH is able to filter larger amounts of critical points than TTC into the bins, but arranges risks in a different order of criticality levels. Besides that, TTC is not precise for the uncomfortable and noticeable bins. RSD can label the existing two types of risk causes into reasonable criticalities. To evaluate 
complete traffic situation risk, we considered in RSD all other TP's within a sensor range of $r\hspace{-0.02cm}=\hspace{-0.02cm}\unit[50]{m}$. Figure \ref{fig:riskmap_rsd_all} shows that RSD filters out correctly the critical TP's on the same path in the front and back. No errors emerge from passing TP's during straight driving 
and turning at intersections. The criticality map has analogous hazard zones at the curve segment ($\text{East}\hspace{-0.04cm}=\hspace{-0.04cm}\unit[180]{m},\text{North}\hspace{-0.04cm}=\hspace{-0.04cm}\unit[400]{m}$). This is achieved with elliptic 2D Gaussians that bend along the curve when using the PMM. 

Remarkably, the red and yellow risk spots in the proximity of intersections are more localized. Overall, we observe that RSD is more selective and has less false positives than the other risk indicators. Since the criticality maps always display the highest detected risk at each spot, and since by construction the same total number of incidents per bin appear in all criticality maps, a smaller area covered by red spots implies that many cases fall onto a single spot. In that sense, the RSD criticality map taking all surrounding cars into consideration is the most specific one in terms of risk localization. A normal distribution around $\mu_{\text{RSD},3}\hspace{-0.02cm}=\hspace{-0.02cm}\unit[0]{m/sec}$ with $\sigma_{\text{RSD},3}\hspace{-0.01cm}=\hspace{-0.01cm}\unit[0.5]{m/sec}$ is observable. RSD rates risks as most critical in $b^*_{1,\text{RSD}}\hspace{-0.05cm}=\hspace{-0.05cm}[1, 0.51]$ when standing in traffic jam with TP's to the sides and in front, while another TP comes from the back with large $-\Delta v$. In the other three bins $b^*_{2,\text{RSD}}\hspace{-0.05cm}=\hspace{-0.05cm}[0.51,0.29]$, $b^*_{3,\text{RSD}}\hspace{-0.05cm}=\hspace{-0.05cm}[0.29,0.11]$ and $b^*_{4,\text{RSD}}\hspace{-0.05cm}=\hspace{-0.05cm}[0.11,0.03]$ this relation is carried on and increases the respective boundary values compared to $b_{2,\text{RSD}}$, $b_{3,\text{RSD}}$ and $b_{4,\text{RSD}}$. It describes the generic notion of higher collision probabilities for denser traffic. In $b^*_{2,\text{RSD}}$ the akin TTC and in $b^*_{3,\text{RSD}}$ plus $b^*_{4,\text{RSD}}$ TH distribution are superposed as in mixture distributions.

\section{Conclusion and Outlook}
\label{sec:outlook}

In this work, we introduced RSD as a new and generalizing risk metric for the criticality assessment of dense and complex traffic situations in real-world operating conditions. For each considered vehicle, RSD first extrapolates kinematic trajectories and finds the overlap between spatial Gaussian distributions to arrive at continuous collision probabilities over future time points. These are then integrated and normalized within an inhomogenous Poisson process of the survival analysis. We improved the robustness of RSD for naturalistic driving with several extensions comprising 2D Gaussians of positional uncertainties (relevant during longitudinal passing), velocity uncertainty (notable e.g.~while waiting at intersections) and the PMM method (for normal distributions that follow sharp curves).

TH and TTC assume a vehicle-to-vehicle crash and characterize the time to the event with different predictions. For TTC the other vehicle has constant speed, whereas TH acts as if it would come to an abrupt halt. In both TH and TTC, the ego vehicle is assumed to continue driving with constant velocity.  
With this in mind, we optimized the velocity uncertainty parameters in RSD so that both the TH and the TTC cases can be incorporated. The parametrization is additionally chosen to comply with sanity rules in car following plus intersection crossing and matched to the statistics in kinematics of the analyzed data. 

An application on NGSIM revealed that RSD is able to differentiate different hazard categories: 1. dynamic stop and go in heavy traffic, 2. approaching with moderate velocity standing front car, and 3. keeping low distance with high speed to the other car. To the contrary, TH as well as TTC are not able to extract the points on the criticality map to the same extent, but rather cover special cases. There are no intense accelerations and accidents in NGSIM. The detected risks of RSD are potential/hypothetical and the scenario always evolved in a way that the cars avoided the criticality. In other words, RSD filters fundamental situations which represent causes for frequent crashes. 

The framework of RSD allows to include not only collision probabilities, but any other type of risk when explicitly modeled (such as disobeying traffic rules, driving off curves or neglecting occlusions). With the help of the survival analysis, RSD steadily outputs a scalar value containing the overall future accident risk of the driving scene at the current time. For this reason, RSD shows to be promising as a standard traffic risk indicator for the validation of AV's. Recording RSD over longer periods in AV's, the driving strategies and errors become ratable and analyzable more easily. To further verify RSD, the criticality map should moreover be compared with heat maps from real accidents for a specific traffic study. Once proven alike, road and traffic sign layouts can be designed to avoid risk spots and to minimize the actual emergence of conflicts.

Behavior planning systems, such as ROPT, balance risk against utility of the travel (i.e., the needed time to arrive at the goal). Since risks with high traffic density are existent, paths which avoid coming close to other cars are automatically preferred. In future work, the averaged risk values at each point of the criticality map could be beneficial as zone risk priors. Consequently, ROPT would even steer slowly through intersections with high risk priors and accelerate in safe segments with low priors. This would correspond to an experienced human driver during his daily commutes. 
At last, a navigation system based on RSD might find routes that are short in distance and low in overall critical event rates. An included ex-post analysis based on RSD and the actually driven trajectories can be calculated in a straightforward way and would make the users' propensity to risk analyzable for each run.

\section*{Acknowledgment}
\noindent This work has been supported by the European Unions Horizon 2020 project \textit{VI-DAS}, under the grant agreement number 690772.




%



\bibliographystyle{IEEEtran}
\bibliography{bib}

\IEEEoverridecommandlockouts

\end{document}